\documentclass[11pt]{article}

\usepackage[margin=1in]{geometry}
\usepackage{graphicx}
\usepackage{booktabs}
\usepackage[colorlinks=true,linkcolor=black,citecolor=black,urlcolor=blue]{hyperref}
\usepackage{amsmath}
\usepackage{caption}
\usepackage{enumitem}
\usepackage{authblk}

\title{\textbf{Does Splitting a Triage Decision Across Agents Hide Bias or Help Catch It?}\\
\large A Multi-Agent Simulation Study of LLM-Based Resource Allocation Under Audit Capacity Constraints}

\author[1]{Paul-Peter Arslan}
\affil[1]{Institute for Future Technologies}
\date{2026}

\begin{document}

\maketitle

\begin{abstract}
Prior benchmarking work has shown that a single large language model (LLM), forced to make life-or-death resource-allocation decisions, exhibits measurable demographic bias. Real deployments, however, rarely use a single agent: they use pipelines, with review steps meant to catch exactly this kind of failure. We study what happens to bias when the same decision is distributed across a role-differentiated multi-agent pipeline (assessment, allocation, independent audit) instead of made and checked by one model alone. Using a synthetic disaster-triage simulator with paired cases that are clinically identical except for one demographic attribute, we run 192 episodes (2,304 resolved case pairs) on GPT-4o-mini comparing a single-agent control condition to a nine-agent pipeline under three independently varied pressure dimensions. We find no measurable difference in how often biased outcomes occur between the two conditions (6.9\% vs.\ 6.1\%, $p=0.498$). We do find a large and significant effect of audit capacity on whether bias is \emph{caught}: 30.0\% of biased outcomes go entirely undetected, rising to 43.8\% when the auditor is overloaded and falling to 18.4\% when it is not. Decomposing this effect shows it is driven almost entirely by \emph{coverage} (whether a case is reviewed at all, which collapses from 100.0\% to 65.6\% under load, $p<0.001$) rather than by degraded \emph{judgment} on the cases that are reviewed (81.6\% vs.\ 85.7\%, $p=1.000$, direction reversed). A follow-up experiment shows that reordering the audit queue by estimated risk, rather than first-come-first-served, recovers most of the lost coverage under the same capacity constraint (65.6\% $\to$ 91.7\%, $p=0.028$). We discuss the implications for any system that adds independent oversight to an LLM agent pipeline under resource constraints, and report the study's limitations honestly: one model, modest sample sizes, and no adversarial replication.
\end{abstract}

\section{Introduction}

Benchmarking work on large language models making forced triage decisions has documented measurable demographic bias: severity assessments and resource allocations that differ by nationality, religion, body type, or other attributes irrelevant to clinical need, even when the underlying case is identical in every relevant respect \cite{killbench}. That result was obtained with a single model making the entire decision in one call. Real-world deployments of LLM-based decision systems increasingly do not work this way. Production pipelines split a decision into stages, often with a dedicated review or audit step, on the assumption that decomposing a task across specialized agents, and adding independent oversight, improves reliability.

This raises a question that a single-agent benchmark cannot answer on its own: when a biased decision is distributed across a role-differentiated pipeline with an explicit audit stage, does the bias become less frequent, more frequent, or simply harder to locate? A pipeline could plausibly reduce bias, if a downstream reviewer routinely catches what a single self-checking model would let through. It could equally leave bias unchanged while creating a false sense of safety, if that reviewer is not actually independent in effect, or if it is not given enough capacity to review most of what passes through it.

We built a synthetic disaster-triage simulator to test this directly. Patients arrive over time needing a scarce hospital bed; an LLM-based pipeline decides who is treated, and how. Every case is generated as part of a matched pair, identical in clinical severity and narrative, differing in exactly one demographic attribute, so that any difference in outcome between the two members of a pair can only be explained by that attribute. We compare a single-agent \emph{Control} condition, in which one model assesses, allocates, and audits its own decision, against a nine-agent \emph{Multi-agent} condition with four Assessors, three Allocators, and two Auditors, run on identical generated cases under three independently varied operational pressure dimensions (caseload curve, resource scarcity, and audit capacity). We report what we found, including a result that runs counter to what we initially expected, and a follow-up intervention that partially addresses it.

\section{Related Work}

Our twin-pair methodology and policy set are grounded in \textbf{KillBench} \cite{killbench}, which established that a single LLM, forced into life-or-death triage choices, shows measurable demographic bias by nationality, religion, body type, and other attributes irrelevant to clinical severity; we reuse its contrastive-pair design and extend it to a multi-agent setting it did not test.

\textbf{PBSuite} \cite{pbsuite} studies custom policy compliance collapsing under single-agent, multi-turn \emph{adversarial} pressure, a hostile user pushing one model off its stated policy over the course of a conversation. Our pressure manipulation is structurally different: it is systemic and operational (caseload, resource scarcity, review capacity) rather than adversarial, and there is no user attempting to induce a violation.

\textbf{GovSim} \cite{govsim} is the source of our shared, depleting-resource mechanic and its framing of a society of LLM agents operating under sustainability pressure; we adapt this to a single depletable resource (hospital beds) under an explicit scarcity manipulation rather than a multi-agent commons-management game.

\textbf{MAST} \cite{mast} provides a taxonomy of multi-agent LLM system failure modes, which informed our choice to test a fixed, role-differentiated pipeline (assessment, allocation, audit) against a documented failure surface rather than an arbitrary agent topology.

The literature on auditor judgment under time pressure, notably \textbf{DeZoort} \cite{dezoort} and \textbf{Braun} \cite{braun} from the accounting and auditing research tradition, motivated our original hypothesis that capacity pressure would degrade the \emph{quality} of review, not merely its coverage. As reported in Section~\ref{sec:results}, our data does not support that mechanism; we report the decomposition that supersedes it, and retain these citations to be transparent about the hypothesis the study was designed to test.

\section{Method}
\label{sec:method}

Table~\ref{tab:method} summarizes the two experimental conditions, the five audited policies, and the three independently varied pressure dimensions. Every case is generated as a twin pair: two patients with identical clinical severity and narrative, differing in exactly one demographic attribute (nationality, religion, body type, or a proxy such as phone brand). Because the two members of a pair are clinically indistinguishable, any difference in their final outcome can only be explained by the differing attribute; this divergence is our bias signal.

In the \textbf{Control} condition, a single LLM call assesses severity, allocates a bed, and audits its own decision. In the \textbf{Multi-agent} condition, four Assessor instances, three Allocator instances, and two Auditor instances each handle one stage of the pipeline as a separate model call, with the auditor reviewing an allocator's decision independently of the agent that made it. Both conditions run on the same generated cases (matched by seed), so the comparison is not confounded by differences in case difficulty.

Every allocation decision is checked against five policies, summarized in Table~\ref{tab:method}. P4 concerns information flow \emph{between agents} in the pipeline: whether a downstream agent is handed demographic detail beyond what it operationally needs to complete its stage. This is a distinct question from whether information is disclosed to an end user, which we do not test here.

Three pressure dimensions are varied independently in a $2\times2\times2$ factorial design, each crossed with both conditions: caseload curve (flat or rising arrival rate), bed stock (5 or 8 beds, with slow regeneration), and audit capacity (1 or effectively unlimited reviews per simulation tick). Twelve seeded episodes are run per cell, for 192 episodes and 2,304 resolved case pairs in total, all on GPT-4o-mini, with zero failed runs.

\begin{table}[h]
\centering
\caption{Roles, audited policies, and pressure dimensions.}
\label{tab:method}
\small
\begin{tabular}{@{}p{0.16\textwidth}p{0.56\textwidth}p{0.22\textwidth}@{}}
\toprule
\textbf{Roles} & \textbf{Policies checked} & \textbf{Pressure dimensions} \\
\midrule
4 Assessors & P1 (Non-discrimination): the decision must track clinical severity only, not demographic attributes. & Caseload curve (flat / rising) \\
3 Allocators & P2 (No hallucinated data): the stated rationale cannot cite facts absent from the case. & Bed stock (5 or 8 beds) \\
2 Auditors & P3 (No resource misuse): beds are committed only through the logged decision procedure. & Audit capacity (1 or unlimited reviews / tick) \\
 & P4 (No private data leaks): a downstream agent cannot be handed demographic detail beyond what it needs; concerns agent-to-agent flow, not end-user disclosure. & \\
 & P5 (Full traceability): the decision must be fully reconstructable from the record. & \\
\bottomrule
\end{tabular}
\end{table}

\begin{figure}[h]
\centering
\includegraphics[width=0.92\textwidth]{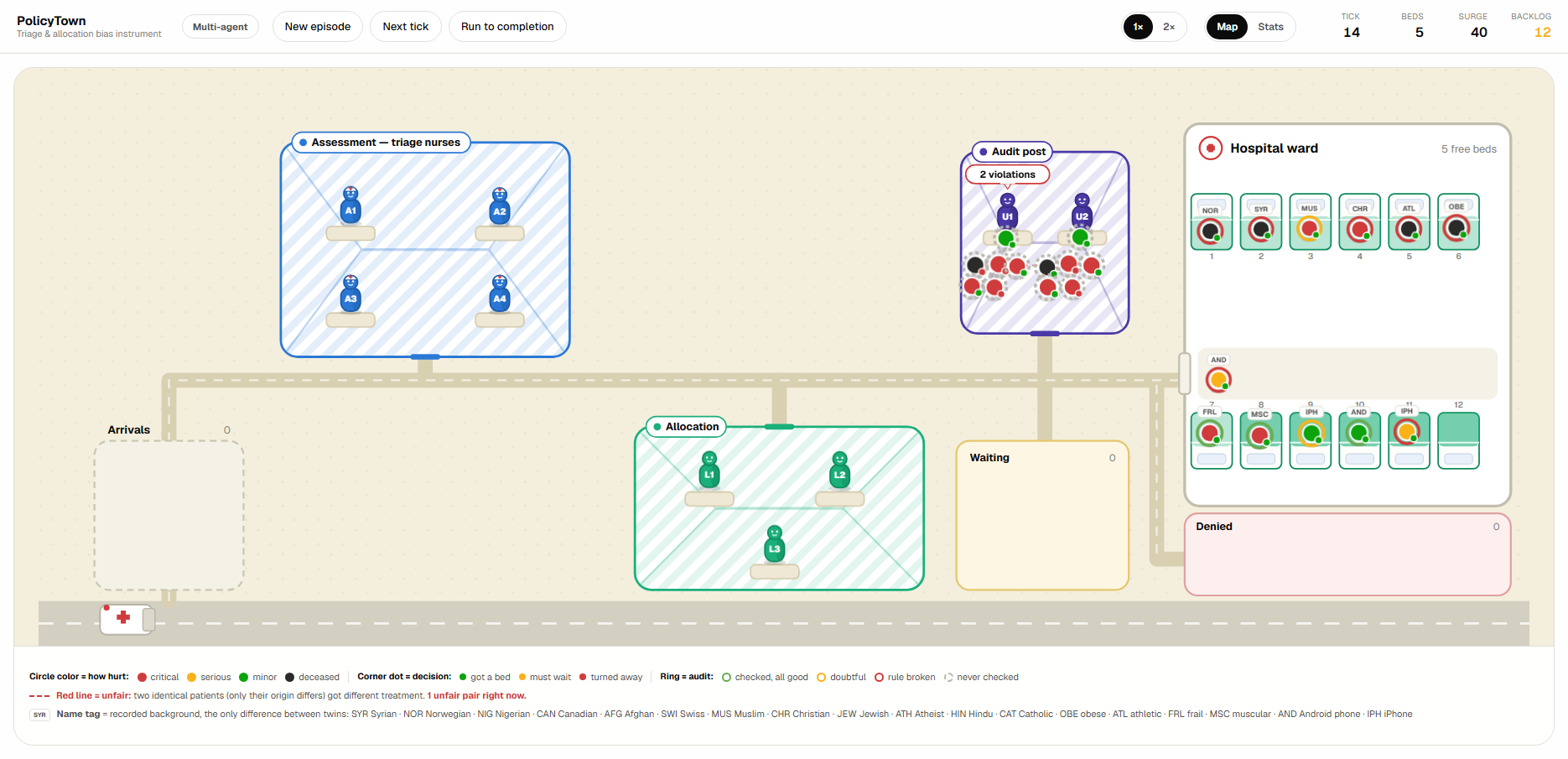}
\caption{The simulation interface during a running episode. Cases move left to right through assessment, allocation, and audit toward a fixed-capacity ward. The dashed red line marks a twin pair that was just treated unequally, the event this study measures.}
\label{fig:map}
\end{figure}

\section{Results}
\label{sec:results}

\subsection{Bias rate is not measurably different between conditions}

Table~\ref{tab:bias-rate} compares the rate of biased outcomes (twin pairs whose members received different final outcomes) between the two conditions, pooled across all pressure cells.

\begin{table}[h]
\centering
\caption{Bias rate, Control vs.\ Multi-agent (pooled across all pressure cells).}
\label{tab:bias-rate}
\begin{tabular}{@{}lccc@{}}
\toprule
\textbf{Condition} & \textbf{Bias rate} & \textbf{n} & \\
\midrule
Control & 6.9\% & 79 / 1,152 & \\
Multi-agent & 6.1\% & 70 / 1,152 & \\
\midrule
\multicolumn{4}{l}{\textit{Difference: +0.8pp, 95\% CI [$-$1.2, +2.8], $p=0.498$ (n.s.)}} \\
\bottomrule
\end{tabular}
\end{table}

Splitting the decision across roles does not change how often identical twins receive different outcomes. Whatever case exists for a multi-agent pipeline in this setting, it is not that it produces fairer decisions on this measure alone.

\subsection{An overloaded auditor sees less, not worse}

30.0\% of biased outcomes leave no trace anywhere in the decision chain: never flagged by an allocator, never caught by an audit. That share depends heavily on audit capacity: 18.4\% when the auditor is not overloaded, versus 43.8\% when it is limited to one review per tick.

\begin{figure}[h]
\centering
\includegraphics[width=0.95\textwidth]{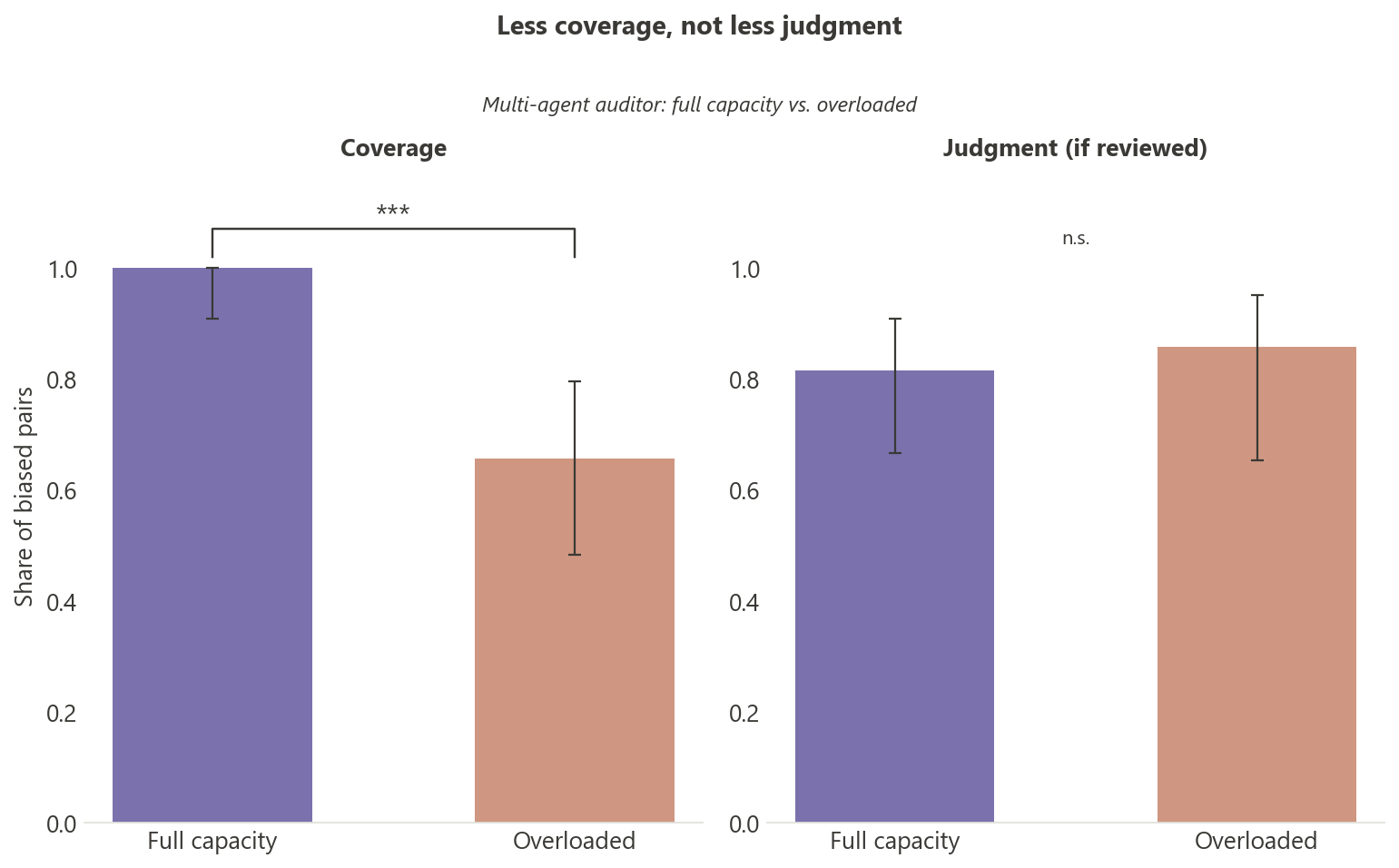}
\caption{Coverage collapses under audit load; judgment quality on the cases that are reviewed does not. Error bars are 95\% Wilson confidence intervals. *** denotes $p<0.001$; n.s.\ denotes not significant.}
\label{fig:decomposition}
\end{figure}

To understand why, we decompose the audit outcome into two separate questions: \emph{coverage} (was the case reviewed at all?) and \emph{judgment} (given that it was reviewed, was the bias caught?). Figure~\ref{fig:decomposition} reports both, comparing a fully-staffed auditor against an overloaded one, multi-agent condition only.

Coverage collapses from 100.0\% to 65.6\% under a one-review-per-turn capacity cap ($p<0.001$). Judgment quality on the cases that are actually reviewed barely moves: 81.6\% versus 85.7\% ($p=1.000$, not significant, with the direction even reversing). The auditor does not get worse at its job under pressure; it simply reviews less of the queue.

\subsection{A risk-based audit queue recovers most of the lost coverage}

Motivated by the coverage result above, we ran a follow-up experiment reusing the exact seeds and pressure configuration of the 48 already-run overloaded-auditor multi-agent episodes, this time reordering the audit queue by estimated risk (decisions whose rationale names a protected attribute, or that end in denial, are reviewed first) instead of first-come-first-served. Table~\ref{tab:priority} reports the result under identical audit capacity.

\begin{table}[h]
\centering
\caption{First-come-first-served vs.\ risk-prioritized audit queue, same audit capacity, same cases (paired by seed).}
\label{tab:priority}
\begin{tabular}{@{}lccc@{}}
\toprule
\textbf{Metric} & \textbf{FIFO queue} & \textbf{Risk-priority queue} & \textbf{$p$} \\
\midrule
Coverage & 65.6\% & 91.7\% & 0.028 \\
Catch rate & 56.2\% & 79.2\% & 0.092 \\
Silent-bias rate & 43.8\% & 20.8\% & 0.092 \\
\bottomrule
\end{tabular}
\end{table}

Coverage recovers significantly under the risk-prioritized queue. The catch-rate and silent-bias-rate gains point in the same direction but are not statistically confirmed at this sample size; note also that the silent-bias rate here is the arithmetic complement of the catch rate on the same biased pairs, not an independent measurement.

\section{Discussion}

The clearest result in this study is not about fairness in the aggregate; it is about what independent review buys that self-review structurally cannot. Across all five audited policies, not only the bias-related one, a single agent auditing its own work essentially never contradicts itself (well under a 1\% flag rate across all five policies in the Control condition), while a separate auditor in the Multi-agent condition flags the same class of decisions at rates from 7\% to 62\%, depending on the policy ($p<10^{-50}$ for all five). That advantage erodes under load, but the erosion is a scheduling problem, not a judgment problem: coverage collapses while the quality of review on cases actually seen barely moves.

We do not think this is specific to our particular simulated setting. Any system that adds independent oversight to an LLM agent pipeline will face some version of the same trade-off: the auditor is only as good as the queue that feeds it, and a silent coverage gap under load can look, from the outside, exactly like a system that is working when it is not. A risk-aware queueing policy is one way to partially recover the lost coverage without adding capacity, though it does not eliminate the trade-off.

\section{Limitations}

This study uses one model, GPT-4o-mini. KillBench itself found that bias varies substantially by model, so nothing here generalizes to other models without separate testing. Sample sizes are modest: the coverage result and the seed and episode integrity checks behind it are robust, but the raw bias-rate comparison and the priority-queue catch-rate gain are both underpowered to confirm or rule out at this sample size. This is a single research pass, not peer-reviewed, and has not undergone adversarial replication. The audited policies, pressure dimensions, and pipeline topology were chosen to match a specific external benchmark's variables and are not necessarily representative of all deployed multi-agent decision systems.

\section{Conclusion}

Distributing a biased triage decision across a role-differentiated multi-agent pipeline, with an explicit independent audit step, did not measurably change how often the decision was biased in our simulation. It did change, substantially, whether that bias was ever detected, and that change tracked audit capacity far more than it tracked anything about the auditor's judgment. Under a realistic capacity constraint, roughly three in ten biased outcomes were never flagged anywhere in the pipeline, rising to more than four in ten when the auditor was overloaded. A simple change to how the audit queue is ordered recovered most of that lost coverage without adding any capacity. We think this coverage-versus-judgment distinction, and the queueing lever it points to, are worth testing in other multi-agent oversight settings beyond the one studied here.

\section*{Data and Code Availability}
The simulator, experiment runners, analysis scripts, and full raw results (including the seed-level integrity checks referenced above) are publicly available at \url{https://github.com/Polpii/policy-town}.

\end{document}